\documentclass[10pt,twocolumn,letterpaper]{article}

\usepackage{tikz}

\newcommand{\blackcircled}[1]{%
  \tikz[baseline=(char.base)]{
    \node[shape=circle, fill=black, text=white, inner sep=1pt] (char) {#1};}%
}

\usepackage{cvpr}              
\usepackage{amsmath}
\usepackage{amssymb}
\usepackage{algorithm}
\usepackage{algorithmic}

\usepackage{booktabs}
\usepackage{comment}
\usepackage{siunitx}
\usepackage{adjustbox}
\usepackage{caption}
\usepackage{placeins}
\usepackage{tikz}
\usetikzlibrary{arrows.meta,positioning,fit,shapes.geometric,calc}

\definecolor{cvprblue}{rgb}{0.21,0.49,0.74}
\usepackage[pagebackref,breaklinks,colorlinks,allcolors=cvprblue]{hyperref}

\def\paperID{17470} 
\def\confName{CVPR}
\def\confYear{2026}

\title{
From Features to Reference Points: Lightweight and Adaptive Fusion for Cooperative Autonomous Driving}

\author{
\textrm{
Yongqi Zhu$^{1}$ \quad
Morui Zhu$^{1}$ \quad
Qi Chen$^{2}$ \quad
Deyuan Qu$^{2}$ \quad
Isabella Luo$^{3}$ \quad
Song Fu$^{1}$ \quad
Qing Yang$^{1}$\thanks{Corresponding Author.}
}\\[5pt]
\textrm{
$^{1}$University of North Texas \quad
$^{2}$Toyota InfoTech Labs
$^{3}$The Hockaday School \quad
}
}

\begin{document}
\maketitle
\begin{abstract}
We present RefPtsFusion, a lightweight and interpretable framework for cooperative autonomous driving. Instead of sharing large feature maps or query embeddings, vehicles exchange compact reference points, e.g, objects' positions, velocities, and size information. This approach shifts the focus from ``what is seen'' to ``where to see,'' creating a sensor- and model-independent framework that works well across vehicles with heterogeneous perception models, while greatly reducing communication bandwidth. To enhance the richness of shared information, we further develop a selective Top-K query fusion that selectively adds high-confidence queries from the sender. It thus achieves a strong balance between accuracy and communication cost. Experiments on the M3CAD dataset show that RefPtsFusion maintains stable perception performance while reducing communication overhead by five orders of magnitude, dropping from hundreds of MB/s to only a few KB/s at 5 FPS (frame per second), compared to traditional feature-level fusion methods. Extensive experiments also demonstrate RefPtsFusion’s strong robustness and consistent transmission behavior, highlighting its potential for scalable, real-time cooperative driving systems.
\vspace{-8mm}
\end{abstract}    

\begin{figure}[!h]
    \centering
    \includegraphics[width=\linewidth]{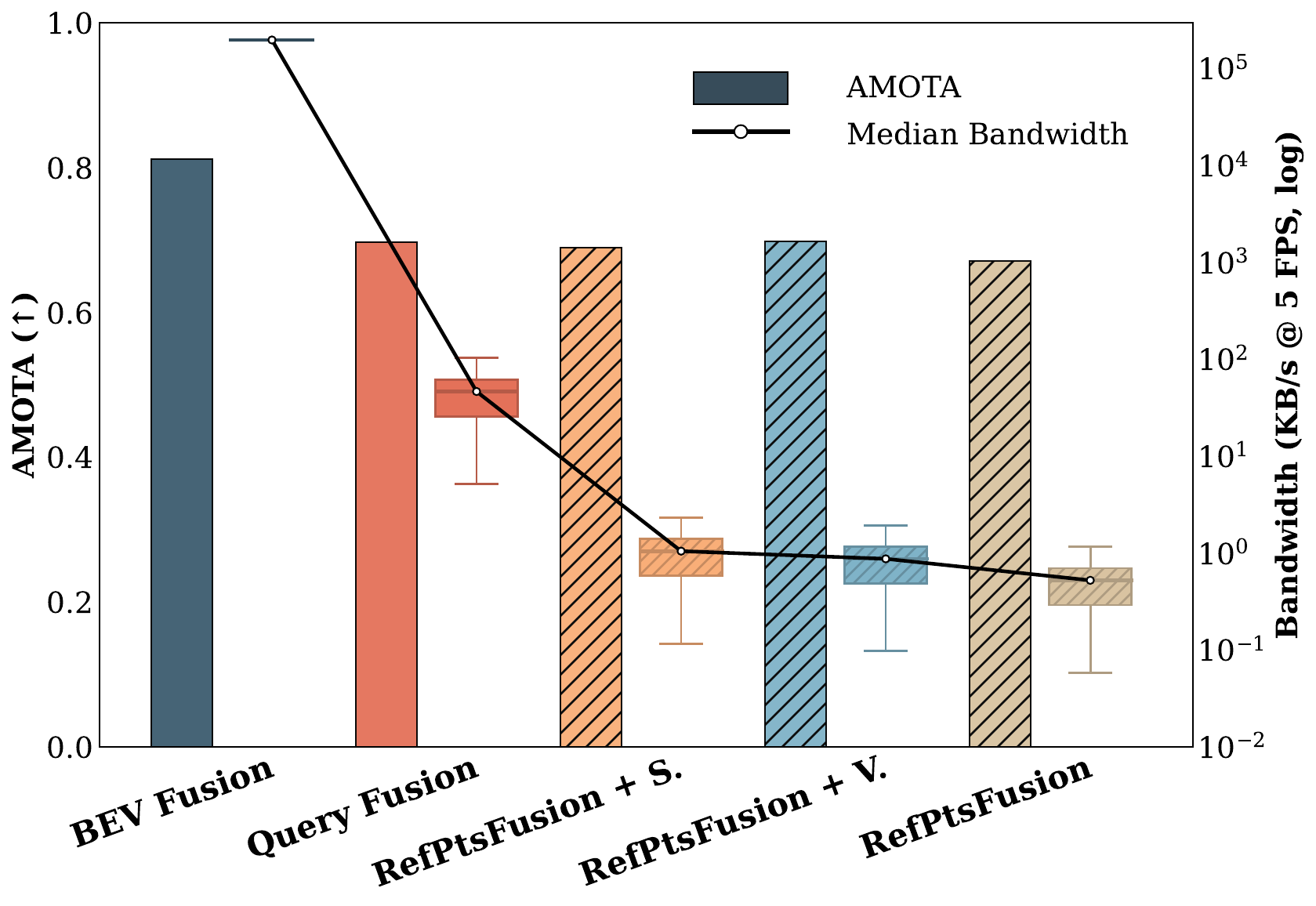}
    \caption{
        \textbf{Performance–Bandwidth Trade-off.}
        Comparison of different cooperative fusion paradigms in terms of perception accuracy (AMOTA) and communication cost.
        Bandwidth is computed using the actual number of effective reference points per frame, reflecting real-time communication.
        The proposed {RefPtsFusion} achieves comparable perception accuracy to feature-level fusion while reducing communication bandwidth by {over five orders of magnitude}.
        %
        Bars with {///} denote methods that explicitly support {heterogeneous model fusion}.
    }
    \label{fig:bandwidth_tradeoff}
    \vspace{-5mm}
\end{figure}
\section{Introduction}
Cooperative autonomous driving (CAD) refers to the coordination and collaboration between multiple autonomous vehicles (AVs), to improve safety, efficiency, and overall performance in a shared driving environment~\cite{automation2020taxonomy}.
Despite its potential, developing reliable CAD systems remains challenging due to the heterogeneity of onboard computing and sensing systems, as well as limitations in current wireless communication networks. 
%
For example, models developed for one type of vehicle may not perform well on others, posing challenges for feature-fusion-based approaches~\cite{head,thornton2025real}, which have recently become popular.
Furthermore, effective cooperation depends on fast and reliable data exchange between vehicles, however, constraints in communication bandwidth would lead to incomplete information sharing, ultimately degrading system performance and safety.
To overcome these challenges, it is crucial to develop a robust cooperative frameworks capable of handling the heterogeneity of vehicle models and adapting to real-time changes in network conditions.

\subsection{Limitations of Prior Work}
Current solutions to CAD can be divided into three categories: high-level, intermediate-level, and low-level data sharing~\cite{han2023collaborative,opv2v}.
High-level sharing (e.g., SAE J2735~\cite{sae}) is bandwidth-efficient but lacks detail and depends heavily on each vehicle’s perception quality. 
%
%
Intermediate-level methods improve accuracy through feature-map fusion~\cite{fcooper,v2vnet}, but require high bandwidth and are suffer from model heterogeneity. 
Low-level sharing of raw sensor data~\cite{cooper} provides richest information, but is prohibitively expensive for real-world deployment. 
%
These limitations motivate a new cooperative paradigm that bridges the gap between high-level and intermediate-level sharing.

\subsection{Proposed Solution}
To bridge this gap, we propose an innovative cooperative framework, namely RefPtsFusion, which achieves both low-bandwidth communication and heterogeneous robust collaboration.
Unlike previous methods, our framework supports both high-level and intermediate-level data fusion, adaptively adjusting the level of shared information based on network conditions. 
At the same time, it effectively handles the heterogeneity among different CAD systems, as shown in Fig.~\ref{fig:bandwidth_tradeoff}.
The key innovation is distinguishing between semantic information and latent features extracted from sensor data by deep neural networks (DNNs). 
For a given perception task, e.g., object detection, a vehicle processes raw sensor inputs through a DNN to generate latent features, which are then converted into semantic outputs. 
Ideally, vehicles can choose to share either semantic or feature data depending on current communication conditions.

To achieve this, we introduce a reference point guided CAD framework designed for heterogeneous autonomous vehicle systems. 
In modern transformer-based vision detection models, {reference points} are spatial indicators that guide the attention mechanism toward specific regions or positions in an image or other types of data. 
They serve as learnable predicted coordinates that help the model understand ``where to see'' when processing raw information. 
%

%
The concept of reference points is widely used in various classic transformer based models, including Deformable DETR~\cite{detr}, Efficient DETR~\cite{yao2021efficient}, BEVFormer~\cite{bevformer}, and the Deformable Attention Transformer (DAT)~\cite{xia2022vision}.
Reference points in transformer architectures show how important they are for helping models understand spatial relationships and interpret complex scenes more effectively.
We discovered that even when vehicles share only their reference points, autonomous driving performance improves significantly compared to non-cooperative approaches.
%
%
Overall, the findings suggest a promising direction: both reference points and latent features play key roles in CAD and can be used individually or together to enhance overall perception performance.

\subsection{Contributions}
Our work makes the following four contributions to the field of cooperative autonomous driving. \blackcircled{1} We propose RefPtsFusion, a novel inter-vehicle collaboration framework that shifts the focus from what is seen (high-dimensional feature sharing) to where to see (low-dimensional, object-level semantics). By exchanging only reference point attributes, including positions, velocities, and sizes, our approach achieves substantial communication savings while remaining sensor- and model-agnostic.
\blackcircled{2} To complement the transmission of purely geometric information, we introduce Selective Top-$K$ Query Fusion, a confidence-aware strategy that selectively integrates a small set of high-confidence queries from sender vehicles. This design enriches geometric reference points with rich features, achieving a strong trade-off between accuracy, robustness, and bandwidth efficiency.
\blackcircled{3} Experimental results demonstrate that RefPtsFusion and its variants maintain stable performance under heterogeneous conditions. The framework effectively filters unreliable objects during fusion, highlighting its robustness and practicality for real-world multi-vehicle systems.
\blackcircled{4} Experiments also show that RefPtsFusion reduces bandwidth consumption by up to five orders of magnitude compared to feature-level fusion methods. 
\section{Related Work}

\textbf{End-to-End Autonomous Driving.} End-to-end autonomous driving (E2EAD) unifies perception, prediction, and planning within a single framework, reducing cascading errors between modules \cite{uniad,genad,stp3}. Recent works leverage BEV representations and multi-task training to jointly address tracking, occupancy prediction, motion forecasting, and planning \cite{bevformer, zhang2025bridging, bench2drive}. While these approaches focus on semantic and geometric consistency, they typically consider single-vehicle settings and overlook challenges in cross-agent communication and bandwidth. In this work, we extend E2EAD to a cooperative setting, proposing a communication-efficient framework for heterogeneous multi-vehicle collaboration.

\textbf{Cooperative Perception.} Cooperative perception enables vehicles to share complementary information to overcome occlusion and limited field-of-view issues \cite{fcooper, cooper, opv2v}. Early fusion exchanges raw sensor data \cite{v2vnet}, achieving complete sensing but requiring high bandwidth. Late fusion reduces communication by sharing detection outputs \cite{han2023collaborative,fadili2025late}, but provides limited information and depends heavily on each vehicle’s perception. Intermediate feature fusion shares BEV features for joint detection \cite{fcooper,cobevt,yazgan2024survey,hu2022where2comm,DiscoNet}, improving occlusion handling yet still incurring substantial bandwidth. Recent works such as mmCooper~\cite{liu2025mmcooper}, ERMVP~\cite{zhang2024ermvp}, and CodeFilling~\cite{hu2024communication} enhance efficiency and robustness, but rely on high-dimensional, task-specific features with limited interpretability.

\textbf{Query Fusion for Cooperative Perception.} Recent studies explore query fusion for cooperative perception, replacing high-dimensional BEV feature sharing with query-level collaboration \cite{univ2x,cooptrack}.
Queries capture rich instance-level semantics, including category, dynamics, trajectory, and interactions, enabling end-to-end cooperative perception with lower bandwidth and better focus on important objects.
However, queries remain high-dimensional continuous embeddings, posing challenges for real-time, large-scale deployment. Additionally, queries from heterogeneous models may carry inconsistent semantics, complicating cross-agent fusion. Finally, query fusion often requires additional end-to-end training, increasing system complexity and deployment cost.


\section{Method}
To tackle the above-mentioned issues, we propose RefPtsFusion, which shifts the focus of inter-vehicle collaboration from ``what is seen'' (high-dimensional features or queries) to ``where to see'' (low-dimensional, physically interpretable object-level semantics). 

\subsection{Overview}
\label{sec:overview}
%
\begin{figure*}[t]
  \centering
  \includegraphics[width=\textwidth]{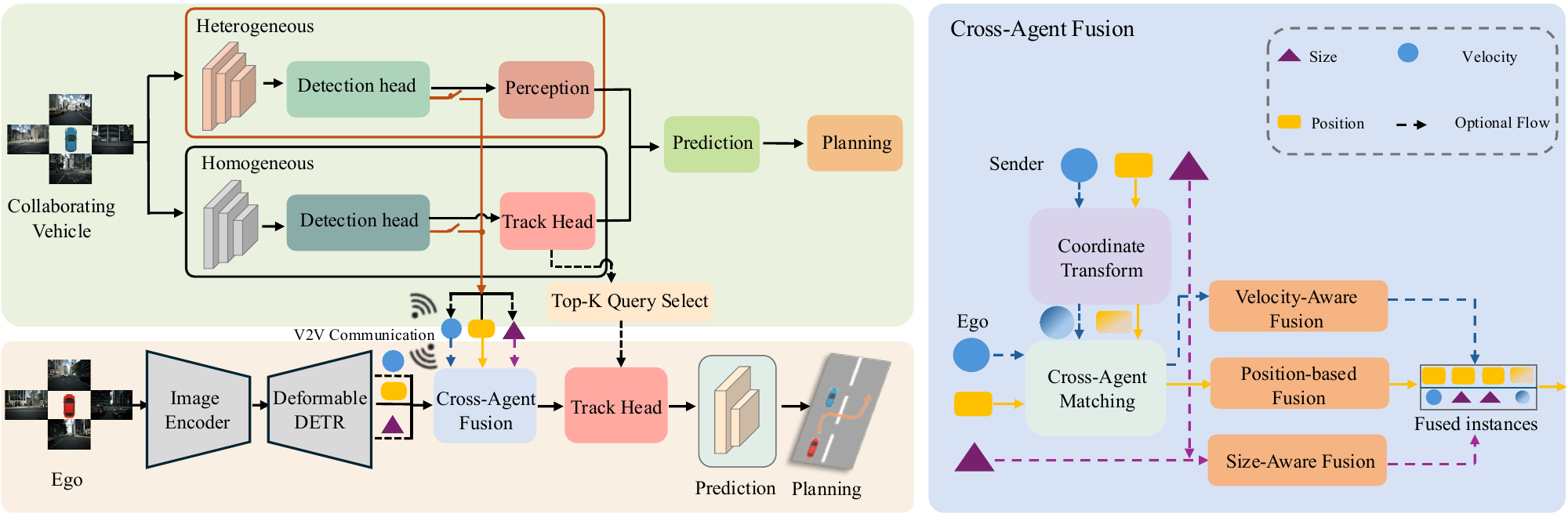}
  \caption{
    \textbf{Overview of the proposed RefPtsFusion framework.}
    It enables cooperative autonomous driving among heterogeneous vehicles through interpretable geometric information. Each sender may employ distinct perception backbones but only needs to transmit reference points, including positions, velocities, and sizes through V2V communication. The ego vehicle performs Cross-Agent Fusion, primarily conducting position-based fusion, while velocity and size information are optionally incorporated, further enhancing downstream perception tasks.
  }
  \vspace{-5mm}
  \label{fig:framework}
\end{figure*}
%
As illustrated in Fig.~\ref{fig:framework}, the proposed RefPtsFusion framework consists of three primary components: the ego vehicle pipeline, the cooperating vehicle pipeline, and the cross-agent RefPtsFusion module.
On the ego vehicle side, we adopt the end-to-end autonomous driving model UniAD as the baseline~\cite{uniad}. After Deformable DETR~\cite{detr} processes the BEV features generated by BEVFormer~\cite{bevformer}, it produces object-level information, including position, size, and velocity. These representations are then fused with the corresponding information shared from the cooperating vehicle through the cross-agent RefPtsFusion module.
The fused outputs are subsequently propagated to downstream modules (e.g., tracking, motion forecasting, and path planning) to accomplish various perception tasks and ultimately generate the final driving trajectory.
We assume that the cooperating vehicle is capable of generating a set of reference points representing the centers of detected objects. 
This assumption is reasonable, as object detection serves as a fundamental capability in most autonomous driving systems.
If the cooperating vehicle additionally provides auxiliary information, e.g., object velocity or size, the RefPtsFusion module can achieve more effective fusion performance.
In this work, we employ Deformable DETR on the cooperating vehicle as an example to validate the effectiveness of the proposed framework; however, the framework itself is model-agnostic and can be seamlessly integrated with any object detection networks adopted by cooperating vehicles.

\subsection{Cross-Agent Fusion}
\label{sec:refpts_fusion}
The core of RefPtsFusion lies in the cross-agent geometric alignment and association of reference points extracted from both the ego and cooperating vehicles. 
The reference points received from the collaborating vehicle are first transformed into the ego vehicle’s coordinate frame, using existing alignment methods (e.g., BB-Alignment~\cite{song2024bb}), the details of which are beyond the scope of this work.
Once aligned, the transformed reference points are matched to their ego-vehicle counterparts, and instance-level fusion is performed to yield a consistent and unified set of fused reference points.
Together with the reference points coordinates, any associated attributes, e.g., velocity and size, will undergo similar transformations and fusion guided by their matched reference points.
We formally describe the fusion process as follows.
Each vehicle $i$ independently detects a set of $N_i$ instances and represents them as
$\mathcal{P}_i = \{ (\mathbf{p}_i^n, \mathbf{v}_i^n, \mathbf{s}_i^n, c_i^n) \}_{n=1}^{N_i}$, where $\mathbf{p}_i^n \in \mathbb{R}^3$ denotes the reference point, $\mathbf{v}_i^n \in \mathbb{R}^2$ the estimated velocity, 
$\mathbf{s}_i^n \in \mathbb{R}^3$ the predicted bounding box size, and $c_i^n$ the confidence score of the $n$-th detected instance. 

Given a sender–ego pair $(S, E)$, the sender's reference points $\mathbf{p}_S^n$ are transformed into the ego coordinate system using the extrinsic calibration~\cite{nuscenes} between vehicles, represented by a transformation matrix $\mathbf{T}_{S \rightarrow E} \in \mathbb{R}^{4 \times 4}$, defined as
\begin{equation}
\mathbf{p}_{S \rightarrow E}^n =
\mathbf{T}_{S \rightarrow E}
\begin{bmatrix}
\mathbf{p}_S^n \\ 1
\end{bmatrix}
=
\begin{bmatrix}
\mathbf{R}_{S \rightarrow E}\mathbf{p}_S^n + \mathbf{t}_{S \rightarrow E} \\
1
\end{bmatrix}.
\label{eq:refpts_transform}
\end{equation}
where $\mathbf{R}_{S \rightarrow E} \in \mathrm{SO}(3)$ and $\mathbf{t}_{S \rightarrow E} \in \mathbb{R}^3$ represent the relative rotation and translation between the sender and ego vehicles in each frame.

After transformation, spatial matching is performed between ego and sender reference points. For each pair $(\mathbf{p}_E^m, \mathbf{p}_{S \rightarrow E}^n)$,
a nearest-neighbor association is established when their Euclidean distance is smaller than a predefined threshold $\tau_d$:
\begin{equation}
\mathcal{M} =
\{ (m, n) \mid
\lVert \mathbf{p}_E^m - \mathbf{p}_{S \rightarrow E}^n \rVert_2 < \tau_d
\}.
\label{eq:match}
\end{equation}
Matched points $\mathcal{M}$ are regarded as the same physical instance and the ego reference points $\mathbf{p}_E^m$ is retained as the canonical representation.
Unmatched sender references are added as new candidates if they fall within the visible range of the ego’s perception field.
The final unified set of reference points is thus expressed as:
\begin{equation}
\mathcal{P}_E^{\text{fused}}
= \mathcal{A}(\mathcal{P}_E, \mathcal{T}(\mathcal{P}_S)),
\label{eq:fused_refpts}
\end{equation}
where $\mathcal{T}(\cdot)$ denotes the geometric transformation in Eq.~\ref{eq:refpts_transform}, and $\mathcal{A}(\cdot)$ represents the deterministic spatial association and aggregation process.
The fused reference points are central to maintaining spatial alignment and temporal consistency across agents. 
They serve as refined spatial anchors for the Tracker module, guiding query updates within both the perception decoder and the Temporal Aggregation Network (TAN)~\cite{motr}.
Serving as precise spatial anchors, each reference point defines the query's location in the current frame and can be propagated or updated according to ego motion or estimated object velocity. 
During temporal attention, they direct the model to focus on relevant regions in subsequent frames, effectively narrowing the search space and improving cross-frame association accuracy.
By fusing reference points instead of raw feature embeddings, the technique achieves explicit geometric alignment across agents and over time, thereby enhancing tracking performance.

\subsection{Velocity- and Size-Aware Enhancements}
\label{sec:velocity_size_refinement}

While fusing reference points captures where to see in each frame, it does not account for temporal dynamics or the spatial extent of objects. 
In tracking, the goal is to maintain consistent identities over time, rather than detecting objects independently in each frame. 
If fusion relies solely on reference point positions, each frame is treated as an isolated instance, i.e., the model knows where an object is at the current moment but cannot predict its motion. 
Rapidly moving, crossing, or occluded objects can easily lead to identity switches or missed associations, as position-only fusion lacks temporal continuity.
To address this limitation, incorporating velocity information is necessary for modeling motion continuity and predicting future positions of reference points. 

In addition, real-world traffic presents significant variations in viewpoints, distances, and occlusions, even among instances of the same vehicle category. 
These factors cause inconsistent bounding box scales and shifted center estimates across agents. 
Integrating object size information helps regularize spatial associations and maintain geometric consistency in multi-agent fusion.
Motivated by these insights, we extend RefPtsFusion with velocity-aware and size-aware enhancements to achieve robust, temporally, and spatially consistent tracking performance.

\textbf{Velocity-Aware Fusion.} In the Tracker module~\cite{motr}, each query maintains an estimated velocity vector, which is used to propagate reference points across frames and constrain cross-frame associations. 
This allows the model to predict the expected location of each instance in the next frame and match it with new detections. 
Integrating velocity information into RefPtsFusion will enable the ego vehicle to preserve temporal continuity after fusion, ensuring more accurate and consistent tracking.

Specifically, for each newly discovered reference point from the cooperating vehicle, we also transform its estimated velocity
$\mathbf{v}_{S \rightarrow E}^n = [v_x, v_y]^\top$ into the ego coordinate frame:
\begin{equation}
\mathbf{v}_{S \rightarrow E}^n =
\mathbf{R}_{S \rightarrow E} \mathbf{v}_S^n.
\label{eq:velocity_transform}
\end{equation}
After transformation, both the position and velocity of the sender’s reference points are integrated into the ego’s fused representation, which is then propagated to the next frame as the initial query state for detection and tracking, allowing the model to update instance locations based on motion priors.
The velocity vector serves two purposes. First, it provides a motion prior for ego queries, enabling the TAN to extrapolate future positions and preserve trajectories even when detections are partially missing or occluded.
Second, it guides temporal attention by focusing the model on regions consistent with object motion, while preventing associations that would imply unrealistic accelerations or abrupt trajectory shifts.
Through this design, velocity-aware fusion enhances temporal consistency and stabilizes long-term object identities in dynamic and partially observable scenarios.


\textbf{Size-Aware Fusion.} To maintain geometric consistency, RefPtsFusion also supports the fusion of object size information. 
Specifically, the predicted sizes from the cooperating vehicle are transformed and aligned with those on the ego vehicle:
\begin{equation}
(\mathbf{p}_{S \rightarrow E}^n,\, \mathbf{s}_{S \rightarrow E}^n)
=
\big(\mathbf{R}_{S \rightarrow E}\mathbf{p}_S^n + \mathbf{t}_{S \rightarrow E},\;
\mathbf{s}_S^n\big).
\label{eq:size_attach}
\end{equation}
%
After geometric alignment and association, the fused object size is stored with the ego query and propagated to the next frame.
Within the perception decoder, object size defines the spatial extent of deformable attention sampling around each reference point, allowing the model to adaptively capture context for targets of varying scales.
During temporal propagation, size provides a geometric prior that constrains cross-frame matching: instances with inconsistent box scales are penalized, whereas those with coherent sizes are favored.
The size information also acts as a geometric constraint that maintains consistent bounding box scales across agents, mitigating mismatched associations caused by partial or biased observations, resulting in more stable identity tracking across time and viewpoints.
%

The combined velocity- and size-aware enhancements elevate RefPtsFusion beyond simple position-based fusion, ensuring both temporal and spatial consistency for reliable cross-agent cooperation.

\subsection{Heterogeneous Backbones}
\label{sec:heterogeneous}
Handling model heterogeneity in cooperative autonomous driving remains a significant challenge. 
In real-world scenarios, vehicles may employ various pipelines built on entirely different backbones, ranging from CNN-based~\cite{center-based,pointpillars} to Transformer-based architectures~\cite{bevformer,true-detr}. 
These models differ not only in their feature representations but also in the structure and semantics of intermediate embeddings. 
Therefore, designing a cooperative framework that works across heterogeneous models is inherently difficult without access to architectural knowledge.

RefPtsFusion addresses this challenge by replacing feature-based fusion with the exchange of interpretable geometric attributes, e.g., position, velocity, and size, commonly available in modern autonomous vehicle's perception systems. 
These attributes form a universal geometric abstraction: position encodes spatial location, velocity captures motion, and size defines spatial extent. 
Each reference point is associated with these attributes and serves as a spatial anchor, linking high-dimensional feature embeddings to explicit locations and guiding where the model attends. 
By converting feature into physically interpretable representations, reference points provide a concise interface that is applicable to any perception backbone architecture. 
Leveraging this principle, RefPtsFusion enables cooperation across heterogeneous models through communication of reference points rather than  features.
%

%
%
When only reference point positions are available, RefPtsFusion performs position-based fusion. 
When velocity or size is provided, these attributes are incorporated to improve temporal consistency and spatial coherence. 
This cooperative process is general, allowing the framework to support a wide range of perception tasks, e.g, detection, tracking, mapping, and occupancy prediction, as long as reference points are used in guiding the perception task.

\subsection{Homogeneous Backbone}
\label{sec:selective_query}
As a general cooperative framework, RefPtsFusion can also be applied to homogeneous backbones. 
When collaborating vehicles use the same perception backbone, cooperative performance can be further enhanced by exchanging high-dimensional features. 
Unlike conventional query- or feature-based fusion methods~\cite{univ2x, cooptrack}, which aggregate all queries or features, we explicitly select the Top-$K$ queries guided by reference points. 
After transforming coordinates into the ego frame, each query from the sender is associated with a confidence score $c_S^n$ produced by the detection head.
A query $\mathbf{q}$ consists of a positional and a semantic part, and only the top-$K$ queries 
$\{\mathbf{q}_S^{(1)}, \dots, \mathbf{q}_S^{(K)}\}$, ranked by their confidence scores, are transmitted by the sender:
\begin{equation}
\mathbf{q} = [\mathbf{q}^{\text{pos}};\, \mathbf{q}^{\text{sem}}], \quad
\mathbf{q}^{\text{pos}}, \mathbf{q}^{\text{sem}} \in \mathbb{R}^{d}.
\label{eq:query_split}
\end{equation}
For each corresponding query pairs on the ego vehicle, we directly update the its semantic embeddings by additive aggregation:
\begin{equation}
\mathbf{q}_{E,\text{sem}}^{(k)*}
=
\mathbf{q}_{E,\text{sem}}^{(k)}
+
\lambda\,\mathbf{q}_{S,\text{sem}}^{(k)},
\quad
k = 1, \dots, K,
\label{eq:add_fusion}
\end{equation}
where $\lambda \in (0, 1]$ is a scaling coefficient.  
The positional components remain unchanged to preserve geometric alignment:
\begin{equation}
\mathbf{q}_{E}^{(k)*}
=
[\mathbf{q}_{E,\text{pos}}^{(k)};\, \mathbf{q}_{E,\text{sem}}^{(k)*}].
\label{eq:final_query}
\end{equation}

This approach can be seamlessly integrated into other cooperative autonomous driving pipelines, offering an interpretable and bandwidth-efficient way to realize feature base fusion.

\section{Experiments}

\subsection{Experimental Setting}

\paragraph{Dataset.}
Our experiments are conducted on the M$^3$CAD~\cite{m3cad}, a large-scale dataset designed to support multi-vehicle end-to-end cooperative autonomous driving research. 
It contains 204 sequences, covering 30,000 frames and over 267K annotated instances across diverse traffic scenarios and challenging weather conditions. 
Most importantly, it supports a broad range of tasks from object detection and tracking to mapping, motion forecasting, occupancy prediction, and planning, thereby facilitating holistic end-to-end evaluation of cooperative driving systems.

\textbf{Evaluation Metrics.} To provide a rigorous and comprehensive assessment, we evaluate performance across five functional modules. 
For tracking, AMOTA, AMOTP~\cite{bernardin2006multiple}, and IDS are used to measure detection accuracy, localization precision, and identity consistency. 
Mapping performance is assessed using IoU-lane and IoU-road to capture topological and geometric fidelity of reconstructed BEV maps. 
Motion forecasting is evaluated with minADE, minFDE, and MR~\cite{liang2020pnpnet, luo2018fast, peri2022forecasting} to quantify trajectory accuracy and miss rate, while occupancy prediction uses IoU to assess spatial accuracy and completeness of predicted occupancy grids~\cite{liu2021swin, zhang2022beverse}. 
For planning, average L2 error evaluates trajectory smoothness, stability, and driving safety~\cite{stp3}.

\textbf{Implementation Details.} We adopt a two stage training and evaluation approached.  
In the first stage, the BEVFormer encoder and perception modules are pretrained for 6 epochs.
The second stage initializes the prediction and planning heads, and fine-tunes the entire network end-to-end for an additional 18 epochs while freezing the pretrained BEVFormer encoder. 
%
%
Training is conducted on 2$\times$ NVIDIA RTX A6000 GPUs.
Additional implementation and hyperparameter details are provided in the supplementary material.

\begin{table*}[!h]
\vspace{-3mm}
    \caption{
Quantitative results on the {M$^3$CAD} dataset demonstrate that the proposed RefPtsFusion and its variants achieve performance comparable to feature- and query-level fusions. 
\textbf{Bold} marks the best performance, while \underline{underline} denotes the second best. 
    }
     \centering
     \resizebox{\textwidth}{!}{
    \begin{tabular}{c|ccc|cc|ccc|cc|c}
    \hline
    \multicolumn{1}{c|}{\textbf{}} & \multicolumn{3}{c|}{\textbf{Tracking}} &\multicolumn{2}{c|}{\textbf{Mapping (\%)}} & \multicolumn{3}{c|}{\textbf{Motion Forecasting}} & \multicolumn{2}{c|}{\textbf{Occupany (\%)}} & \multicolumn{1}{c}{\textbf{Planning ($m$)}}\\
    \textbf{Methods} & {AMOTA$\uparrow$} & {AMOTP$\downarrow$} & {Recall$\uparrow$} & {IoU-R.$\uparrow$} & {IoU-L.$\uparrow$} & {ADE$\downarrow$} & {FDE$\downarrow$} & {MR$\downarrow$} & {IoU-n$\uparrow$} & {IoU-f$\uparrow$} & {L2$\downarrow$}\\ 
    \hline
    No fusion~\cite{uniad} & 0.254 & \underline{0.495} & 0.455 & 94.3 & 53.8 & \underline{0.312} & \underline{0.336} & \underline{0.003} & 78.8 & \underline{65.4} & 0.401 \\
    F-Cooper~\cite{fcooper} & \underline{0.720} & 0.680 & 0.816 & - & - & - & - & - & - & - & -\\
    M$^3$CAD~\cite{m3cad} & \textbf{0.812} & \textbf{0.490} & \textbf{0.916} & \underline{95.7} & \underline{56.7} & \textbf{0.251} & \textbf{0.262} & \textbf{0.001} & \textbf{86.2} & \textbf{73.3} & {0.234} \\
    UniV2X~\cite{univ2x} & 0.697 & 0.601 & \underline{0.835} & 95.6 & 55.6 & 0.349 & 0.363 & \textbf{0.001} & \underline{80.5} & 63.8 & \textbf{0.221} \\
    RefPtsFusion (Ours) & 0.671 & 0.684 & 0.758 & \textbf{96.0} & \textbf{58.3} & 0.346 & 0.358 & \textbf{0.001} & 79.5 & 62.5 & 0.300 \\
    RefPtsFusion + V. & 0.698 & 0.649 & 0.802 & \textbf{96.0} & \textbf{58.3} & 0.349 & 0.364 & \textbf{0.001} & 79.4 & 62.3 & 0.307 \\
    RefPtsFusion + S. & 0.690 & 0.707 & 0.800 & \textbf{96.0} & \textbf{58.3} & 0.362 & 0.376 & \textbf{0.001} & {79.5} & 62.6 & \underline{0.232} \\
    \hline
    \end{tabular}}
    \label{tab:quantitative_main}
\end{table*}

\subsection{Quantitative Comparison}
We compare our method against several representative baselines: No Fusion, an ego vehicle model without cooperation; F-Cooper~\cite{fcooper}, the first feature-level cooperative perception approach; M$^3$CAD~\cite{m3cad}, the most recent end-to-end feature-level fusion framework; and UniV2X~\cite{univ2x}, a query-level method that exchanges high-dimensional queries across vehicles.

Table~\ref{tab:quantitative_main} shows that RefPtsFusion achieves competitive perception and planning performance by sharing only reference points among vehicles.
Compared with feature- and query-based fusion methods, RefPtsFusion achieves competitive tracking accuracy (0.671 AMOTA and 0.758 Recall), only 0.037 and 0.092 lower than Query Fusion (0.697 AMOTA and 0.835 Recall).
It also performs comparably on other tasks, e.g., motion forecasting and occupancy prediction, and offers slightly better results on mapping accuracy (96.0\% IoU-Road, 58.3\% IoU-Lane). 
This may be because, during the first-stage training, TrackFormer (in UniAD~\cite{uniad}) was optimized using fused reference points, which provided additional supervisory signals to adjust the parameters of BEVFormer. 
As a result, the generated BEV features became more informative and structured, offering richer spatial cues that enhanced MapFormer’s performance on the mapping task.

These results demonstrate that explicit fusing reference points is useful in capturing cross-agent correlations for effective cooperative perception.
Incorporating velocity information (RefPtsFusion + V.) further enhances temporal consistency in tracking, increasing AMOTA to 0.698 and Recall to 0.802, achieving performance nearly on par with UniV2X. 
This implies reference points information can effectively improve motion alignment across agents without relying on high-dimensional features.
Overall, RefPtsFusion delivers accuracy comparable to feature- and query-based fusion methods, while maintaining an interpretable and lightweight design, highlighting its effectiveness for end-to-end cooperative autonomous driving.

\subsection{Simulate Heterogeneous Backbones}
\label{sec:heterogeneous simulation}

To evaluate how RefPtsFusion effectively fuses reference points from heterogeneous sender modules, we simulate sender's object detection results with varying qualities by injecting controlled levels of false negatives (FN) and false positives (FP) based on the ground truth (GT). 
This setup reflects real-world autonomous driving scenarios, where sender vehicles may have differing detection accuracy due to model heterogeneity, as well as the variations in sensor configurations.

\begin{table}[!t]
  \caption{
      Evaluation under heterogeneous sender simulation. }
  \label{tab:heterogeneous_sender}
  \centering
  \setlength{\tabcolsep}{2.8pt}
  \renewcommand{\arraystretch}{1.05}
  \resizebox{0.98\linewidth}{!}{
  \begin{tabular}{lcccccc}
    \toprule
    Method & AMOTA $\uparrow$ & AMOTP $\downarrow$ & RECALL $\uparrow$ \\
    \midrule
    RefPtsFusion & 0.671 & 0.684 & 0.758 \\
    \hline
    RefPtsFusion + GT & 0.706 & 0.647 & 0.806 \\
    \hline
    FN 10\%, FP 0\% & 0.706 & 0.648 & 0.780 \\
    FN 20\%, FP 0\% & 0.709 & 0.656 & 0.782 \\
    FN 40\%, FP 0\% & 0.690 & 0.699 & 0.780 \\
    FN 60\%, FP 0\% & 0.697 & 0.700 & 0.759 \\
    FN 0\%, FP 10\% & 0.693 & 0.702 & 0.830 \\
    FN 0\%, FP 30\% & 0.692 & 0.702 & 0.761 \\
    FN 0\%, FP 50\% & 0.692 & 0.701 & 0.784 \\
    \bottomrule
  \end{tabular}
  }
\end{table}

As shown in Table~\ref{tab:heterogeneous_sender}, RefPtsFusion achieves stable performance across a wide spectrum of sender detection qualities.  
When false negatives increase from 10\% to 60\% while false positives remain at zero, AMOTA remains within a narrow range (0.706–0.697) and recall fluctuates slightly (0.780–0.759), indicating that the framework can effectively leverage partial and incomplete sender information.  
Similarly, when false positives increases (from 10\% to 50\%), RefPtsFusion still achieves strong results, reaching the  competitive AMOTA (0.692-0.693) and recall (0.761-0.830). 
%
These results suggest that RefPtsFusion can accommodate senders with varying models and reliabilities, enabling robust performance in heterogeneous cooperative driving systems.

\subsection{Selective Top-K Query Fusion}
\label{sec:select query}

%
In a homogeneous system setup, sender queries are ranked according to the confidence scores of their corresponding reference points. Only the top-$K$ queries are transmitted from the sender. 
We vary $K$ from 5 to 100 to analyze the trade-off between semantic enhancement and feature interference introduced by query fusion.

\begin{table}[!t]
  \caption{
  Impact of the number of fused queries ($K$) on AMOTA and Recall in the proposed RefPtsFusion method.}
  \label{tab:query_fusion_ablation}
  \centering
  \setlength{\tabcolsep}{2.5pt} 
  \renewcommand{\arraystretch}{1.1} 
  \resizebox{0.95\linewidth}{!}{ 
  \begin{tabular}{lcc}
    \toprule
    Method & AMOTA $\uparrow$ & Recall $\uparrow$ \\
    \midrule
    UniV2X~\cite{univ2x} & 0.697 & 0.835 \\
    \hline
    RefPtsFusion & 0.671 & 0.758 \\
    RefPtsFusion + 5 Queries & 0.696 & 0.797 \\
    RefPtsFusion + 10 Queries & {0.698} & {0.802} \\
    RefPtsFusion + 50 Queries & 0.694 & 0.779 \\
    RefPtsFusion + 100 Queries & 0.689 & 0.776 \\
    \bottomrule
  \end{tabular}
  }
\end{table}

\begin{figure}[t]
    \centering
    \includegraphics[width=\linewidth]{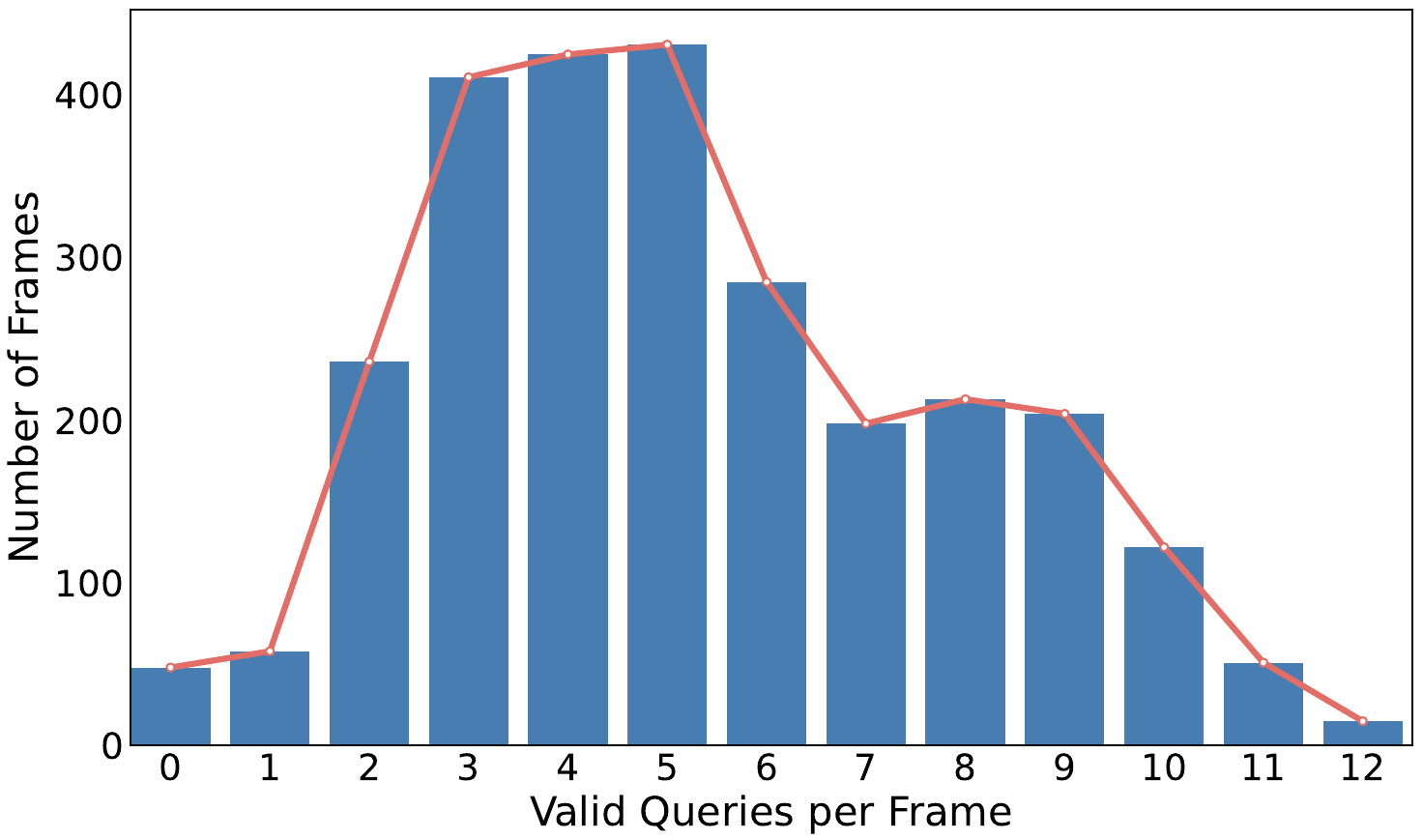}
    \caption{
        {Distribution of valid queries per frame.}
    }
    \vspace{-5mm}
    \label{fig:distribution}
\end{figure}

As shown in Table~\ref{tab:query_fusion_ablation}, selectively fusing a small number of sender queries leads to an improvement over the basic RefPtsFusion. 
Fusing only the top-10 queries increases AMOTA from 0.671 to 0.698 and recall from 0.758 to 0.802, matching the performance of the UniV2X baseline (0.697).
%
However, when $K$ exceeds 10, performance begins to degrade as additional queries contribute little meaningful information to the perception task. 
This trend suggests that only a limited subset of query features effectively complements the geometric representation of the reference points on the ego vehicle.
As shown in Figure~\ref{fig:distribution}, among the queries transmitted from the collaborating vehicle, only an average of 5.27 correspond to ground-truth objects. 
When the sender selects the top 10 queries, however, nearly all reference points matching the ground truth ($>97.6\%$) are successfully shared with the ego vehicle.
Overall, these results confirm that the proposed selective top-$K$ query fusion achieves strong cooperative perception performance while preserving the interpretable nature of the RefPtsFusion framework.

\begin{figure*}[t]
    \centering
    \includegraphics[width=1\linewidth]{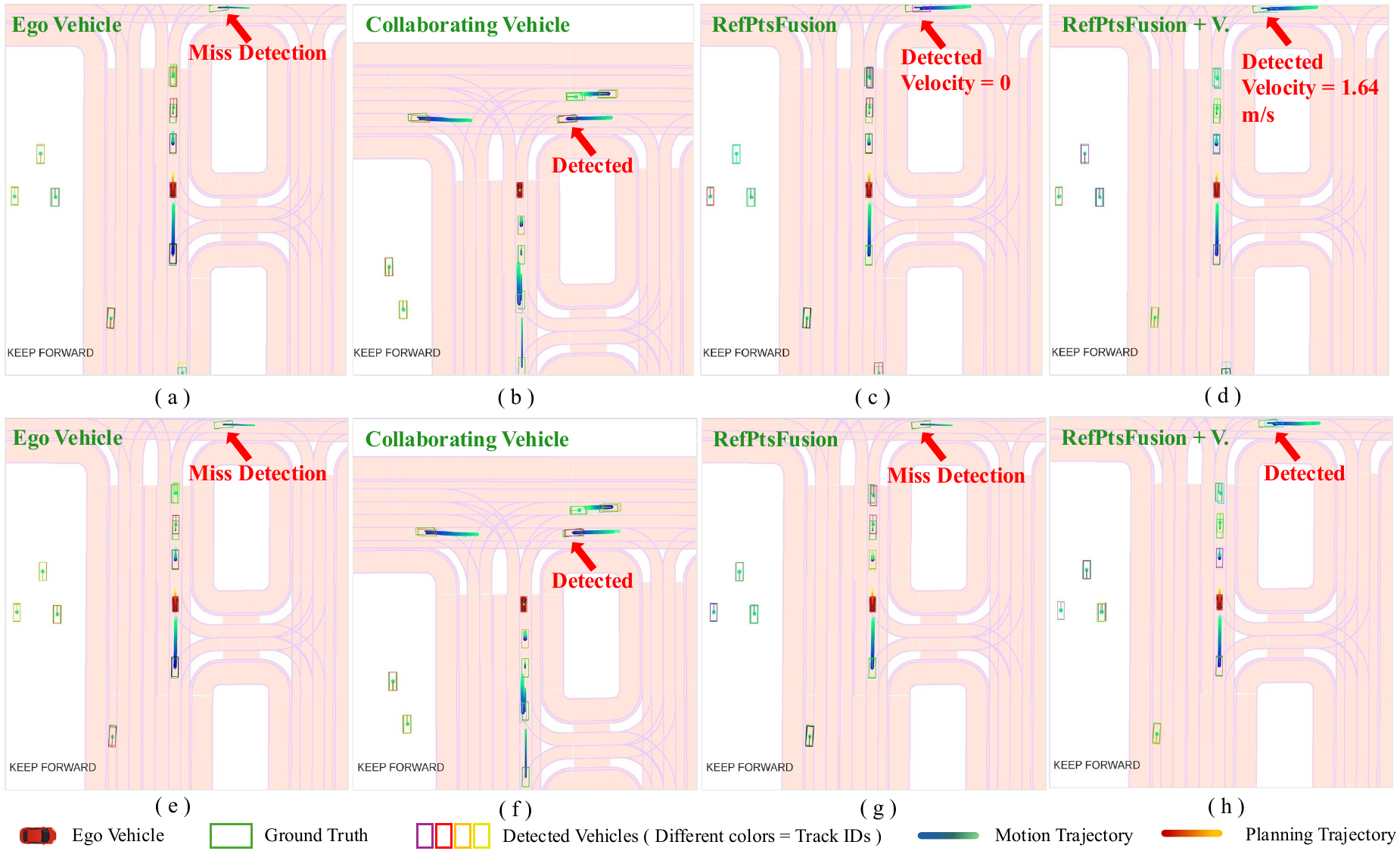}
    \caption{
Qualitative comparison of cooperative perception over two consecutive frames. At time $t_n$, the ego vehicle fails to detect an object (a), while it is successfully perceived by the sender (b). With reference point fusion, both RefPtsFusion (c) and RefPtsFusion + V. (d) correctly localize the object. At the next frame $t_{n+1}$, the detection from RefPtsFusion gradually fades or disappears (g), whereas RefPtsFusion + V. maintains a stable detection (h), highlighting the benefit of incorporating velocity cues for temporal consistency. 
    }
    \vspace{-5mm}
    \label{fig:qualitative_speed}
\end{figure*}   

\subsection{Communication Cost Analysis}
\label{sec:communication cost analysis}
To understand the network bandwidth requirements of different cooperative solutions, we investigate the maximium amount data shared from a collaborating vehicle with a data rate of 5 FPS (frame per second). 
As shown in Table~\ref{tab:bandwidth_summary}, traditional feature- and query-level cooperative perception methods, e.g., M$^3$CAD and UniV2X, demand extremely high communication bandwidth due to the transmission of dense BEV features or high-dimensional query embeddings.
We observe up to $200,000$~KB/s and $4,693$~KB/s bandwidth required for M$^3$CAD and UniV2X, respectively. 
In contrast, RefPtsFusion based solutions dramatically reduces the requirement by exchanging only geometric information, such as object position, velocity, and size. 
Even with all attributes included, the total bandwidth usage remains below 141~KB/s, achieving over $10^3\times$ reduction compared to conventional approaches. 
Note that the actual data transmitted in RefPtsFusion is much smaller than the maximum payload required to send all 900 reference points. 
In practice, only a small subset of (around 10) queries is transmitted, reducing the average bandwidth to 3 KB/s, corresponding to a $10^5\times$ reduction in network traffic.

\begin{table}[!h]
    \caption{
  Quantitative comparison of communication payloads and bandwidth requirements at 5~FPS.
  %
  }
  \label{tab:bandwidth_summary}
  \centering
  \setlength{\tabcolsep}{6pt} 
  \renewcommand{\arraystretch}{1.05} 
  \resizebox{\linewidth}{!}{ 
  \begin{tabular}{lcc}
    \toprule
    \textbf{Fusion Methods} & \textbf{Max. Payload / Frame} & \textbf{Bandwidth} \\
    \midrule
    M$^3$CAD & 40{,}000~KB & 200{,}000~KB/s \\
    UniV2X & 939~KB & 4{,}693~KB/s \\
    RefPtsFusion & 10.5~KB & 52.7~KB/s \\
    RefPtsFusion + V. & 17.6~KB & 87.9~KB/s \\
    RefPtsFusion + S. & 21.1~KB & 105.5~KB/s \\
    RefPtsFusion + V.S. & 28.2~KB & 141.0~KB/s \\
    \bottomrule
  \end{tabular}
  }
\end{table}

\subsection{Qualitative Analysis}
To further demonstrate the effectiveness of the proposed RefPtsFusion, we visualize representative qualitative results in Fig.~\ref{fig:qualitative_speed}.
This visualization focuses on the comparison between the base {RefPtsFusion} model and the {RefPtsFusion + V.} variant, highlighting the benefits of integrating speed information.
Additional qualitative results for other variants are provided in the supplementary material.


As shown in the top-row subfigures of Fig.~\ref{fig:qualitative_speed}, at time $t_n$, the ego vehicle fails to detect an object (in Fig.~\ref{fig:qualitative_speed}(a)), whereas it is successfully perceived by the sender (in Fig.~\ref{fig:qualitative_speed}(b)). 
With reference point fusion, both the {RefPtsFusion} and {RefPtsFusion + V.} variants correctly localize the object (in Fig.~\ref{fig:qualitative_speed}(c) and Fig.~\ref{fig:qualitative_speed}(d)).
At the next frame $t_{n+1}$, the detection obtained by {RefPtsFusion} gradually fades or disappears (in Fig.~\ref{fig:qualitative_speed}(g)), while the {RefPtsFusion + V.} variant maintains a stable detection for the same object (in Fig.~\ref{fig:qualitative_speed}(h)). 
This highlights a key limitation of reference point–only fusion: although the location of a mis-detected object can be transferred from the collaborating vehicle, the absence of motion context makes it difficult to maintain the object’s temporal identity, often leading to premature disappearance or filtering in subsequent frames.

By incorporating a velocity prior, the {RefPtsFusion + V.} variant not only enhances temporal tracking but also improves the ``survivability'' of newly introduced objects. 
These results suggest that velocity cues provide valuable dynamic information, allowing the detection head to reinforce confidence in true positives, thereby reducing false negatives and improving temporal consistency across frames.

\section{Conclusions}
In this work, we introduced RefPtsFusion, a lightweight and interpretable framework for heterogeneous cooperative autonomous driving. 
By exchanging only semantic information, our approach reduces bandwidth consumption by over five orders of magnitude compared to feature-level fusion, while maintaining comparable performance. 
Leveraging reference points also enables the selection of high-quality features or queries, making traditional feature- or query-based fusion more network-efficient.
{Future work} will explore extending RefPtsFusion beyond the tracking function, investigating how to implement it in other perception tasks, e.g., cooperative occupancy prediction and motion planning. 
%

%




{
    \small
    \bibliographystyle{ieeenat_fullname}
    \bibliography{main}
}


\end{document}